\documentclass[final,3p,times]{elsarticle}
\usepackage{amssymb}
\usepackage{amsthm}
\usepackage{hyperref}
\usepackage{flushend}
\usepackage[utf8]{inputenc}
\usepackage{graphicx} 
\usepackage{subfigure} 
\usepackage{color,soul}
\usepackage{cancel}
\usepackage{etoolbox}
\makeatletter
\patchcmd{\ps@pprintTitle}
  {Elsevier}
  {Renewable Energy.}
  {}{}

\journal{}

\newcommand{\rfig}[1]{Fig.~\ref{#1}}

\newcommand{\requ}[1]{Eq.~\ref{#1}}

\begin{document}

\begin{frontmatter}



\title{New approach for solar tracking systems based on computer vision, low cost hardware and deep learning}


   \address[add1]{CIEMAT- Plataforma Solar de Almería, Ctra. de Senés s/n Tabernas, 04200 Almería, Spain.}
  \address[add2]{CIESOL Research Centre for Solar Energy, UAL-PSA.CIEMAT Joint Centre, Almería,Spain.}
   \address[add4]{University of Almería, Ctra. Sacramento s/n, Almería 04120, Spain.}

  \author[add1,add2]{Jose A. Carballo}
  \ead{jcarballo@psa.es}
  \author[add1,add2]{Javier Bonilla}
  \ead{javier.bonilla@psa.es}
  \author[add2,add4]{Manuel Berenguel}
  \ead{beren@ual.es}
    \author[add1]{Jesús Fernández-Reche}
  \ead{jesus.fernandez@psa.es}
    \author[add1]{Ginés García}
  \ead{gines.garcia@psa.es}




\begin{abstract}

In this work, a new approach for Sun tracking systems is presented. Due to the current system limitations regarding costs and operational problems, a new approach based on low cost, computer vision open hardware and deep learning has been developed. The preliminary tests carried out successfully in Plataforma solar de Almería (\emph{PSA}), reveal the great potential and show the new approach as a good alternative to traditional systems. The proposed approach can provide key variables for the Sun tracking system control like cloud movements prediction, block and shadow detection, atmospheric attenuation or measures of concentrated solar radiation,  which can improve the control strategies of the system and therefore the system performance.

\end{abstract}


\end{frontmatter}


\section{Introduction}

Industrial evolution and population growth, have motivated the continuous increase of the world energy demand. The perspectives indicate that, world population will continue to increase, specially for non Organisation for Economic Co-operation and Development (\emph{OECD}) countries demanding more energy for residential uses \cite{UNDESA2017}. Furthermore,  energy consumption due to the  industrial sector, which continues to account the largest share of the energy consumption, is also expected to increase. For all that, it is predicted that World energy consumption will rise 28 \% between 2015 and 2040 \cite{EIA2017}.

In the past, the demand has been covered by traditional energy sources such as fossil and nuclear fuels which heavily pollute the ecosystem. The consequences are being revealed now \cite{Kannan2016}. For that, it is necessary to change the current energetic model to mitigate the environmental issues and reply to new demands in a clean and sustainable way. Recent studies indicate that 100 \% of renewable electricity and energy in general supply can be achieved in the top industrialized countries. These studies show how the transition should be \cite{Hohmeyer2015,Jacobson2015}, its problems and its economic, environmental and social benefits \cite{scheer2013solar}.

These studies suggest that all kind of renewable energies sources should be employed, creating a sustainable and reliable energy mix. Note that the development carried out in different renewable energies sources (solar, wind, hydropower and geothermal) makes them  viable sources of energy, even in places rich in fuel reserves. The contribution in the world energetic model of the renewables energies is the fastest growing \cite{EIA2017}.

Although several renewable technologies are already competitive, researchers are undertaking investigations in order to increase efficiency and reduce cost \cite{Hohmeyer2015}, as shown by the amount of scientific production related to it \cite{Sheikh2016}.

Among all renewable technologies, those that take direct advantage of solar radiation flux have lot of potential and they generated many different subcategories. Most of them improve the efficiency concentrating the solar flux by optical systems or exposing the receiver so it captures as much solar radiation as possible. 

\subsection{Sun tracking}

Solar energy technologies convert the solar radiation into other type of energy, therefore estimating the availability and the nature of the source of energy is a key issue. The most important characteristic of the solar radiation is that the relative position of the Sun in the sky is constantly changing generating daily and annual cycles due to the Earth rotation around its axis and translation around the Sun. This fact causes that the systems  do not receive all the radiation in an optimal way for a fixed position. For that, the solar energy collectors in energy generation systems demand a solar tracking system (\emph{STS}) to control the alignment with the Sun to increase the received solar radiation. Furthermore, the \emph{STS} is responsible for managing the basic tasks that guarantee the correct daily operation. Among these tasks the most relevant are: to compute the tracking setpoint, communications management, diagnosis of faults or errors, to control drive mechanism and the decision making in emergency situations.

In the literature there are some recent reviews about \emph{STSs} that show the growing interest in these systems. For instance, Lee et all. {\cite{Lee2009}} presented a review of the major algorithms for sun tracking systems developed over the past 20 years and concluded confirming the great applicability of sun tracking system for a diverse range of high-performance solar-based applications. Prinsloo and Dobson {\cite{SolarTracking2015}} carried out an extensive review about computer-based sun tracking devices, high-precision solar position algorithms, software for computing the solar vector, solar coordinates and sun angles, showing the potential and rapid growth in recent years. Later, Mousazadeh et all. {\cite{Mousazadeh2009}} performed an extensive review and discussed pros and cons of each system. Recently, Nsengiyumva et all. {\cite{Nsengiyumva2018}} classified the current \emph{STSs} and stated that closing the control loop in \emph{STSs} could significantly improve the overall system performance.

Nowadays, \emph{STSs} can be classified according to the definition method of the solar relative position in the sky. There are two main methods, solar position algorithms or optical methods/algorithms. Furthermore, \emph{STSs} can be classified regarding the number and position of rotation axes in one or two rotation axes mainly. For one rotation axis, horizontal, vertical and tilted-axis systems can be found as subcategories. Azimuth-elevation and polar trackers can be found as subcategories for two rotation axes trackers. 

Sun trackers can guide the system with continuous or discrete movement depending on the rotation system and the controller. Regarding to the control of movements, \emph{STSs} can also be classified  as passive and active controllers. Passive solar trackers usually  are composed by a couple of actuators working against each other. Passive trackers are based on thermal expansion and using the different solar radiation conditions orient the system in the direction where the radiation over both actuators is the same \cite{Mousazadeh2009}. Active trackers can be classified in microprocessor and electro-optical sensor based, computer controller date and time based, auxiliary bifacial solar cell based and a combination of these three systems \cite{Mousazadeh2009}. Also, active solar trackers can be classified in closed-loop and open-loop controllers. Closed-loop controllers  are based on feedback transferred from sensors which identify parameters induced by the environment. On the other hand, an open-loop controller, estimates its inputs using only the current state, without using feedback to determine if its inputs have achieved the desired goal, so this one is simpler than the closed-loop controller but it cannot correct any errors and may not counteract for disturbances in the system \cite{Lee2009}. Finally, it can be found centralized or distributed \emph{STS} controllers in large plants dedicated to the production of energy. Note that most of \emph{STSs} are developed for a specific solar technology or purpose \cite{berenguel2012advanced}.

Among all the errors that can affect a solar energy collector \cite{Chiesi2017}, \emph{aiming errors}, \emph{tracking offset} and optical errors are those directly related to the \emph{STS}. The first one is related to constant aberrations such as, pedestal tilt due to displacement of the center of gravity, bad reference or structure deformations due to wind loads. Also, the effect of the refraction of solar radiation in the atmosphere can be classified as an aiming error because it can modify the apparent position of the Sun due to the conditions of the atmosphere \cite{Jenkins2013}. The \emph{tracking offset} is related to the \emph{STS} abilities in general (movement, controller and tracker resolution). The last of the three is related to optical quality and it can be found astigmatic aberration as the most connected with the \emph{STS}. Although astigmatic aberration is not caused by \emph{STS} directly, correct control strategies of \emph{STSs} can help to reduce its effects (spillage and decrease concentration factor).

With respect to active control solar trackers, computer controlled based on date and time, that are the most employed,  gets the solar position according to solar equations that need time, date and location as inputs. The accuracy of this system is very conditioned by constant aberrations. On the other hand, bifacial solar cell and electro-optical sensor systems are time and location independent although they are only able to obtain the relative solar position and control solar systems when the system optical axis should be aligned with the Sun vector, for example parabolic dishes or photovoltaic (PV) two axes trackers. El Kadmiri et al. \cite{ElKadmiri2015} developed a solar tracker based on omnidirectional computer vision able to extract accurate information about the Sun position and Ruelas et al. \cite{Ruelas2013} developed a system based on a video processing sensor able to determine the Sun position using low cost devices.  The accuracy of these systems is mostly influenced by the controller and sensor resolution.  All of them demand a specific individual configuration during the first start-up of the system. 

The new approach for \emph{STSs} proposed in this work is based on the use of computer vision techniques to carry out the Sun tracking task,  furthermore  the proposed system enables computing some key variables related to this. Particularly, the new approach makes use of computer vision techniques related to object detection with region proposal techniques based on deep learning by means of convolutional neural networks (\emph{CNN}). In this work the implementation of the approach in a real system is analyzed and discussed.

The authors of the present work have patented the Sun tracking approach based on computer vision presented \cite{carballoPatent}. Furthermore, they also presented a small prototype aims at educational purposes \cite{Carballo}. 

\section{Solar tracking system based on computer vision}

\begin{figure}[h!]
   \centering
 \includegraphics[angle =-90, width = 0.72\columnwidth]{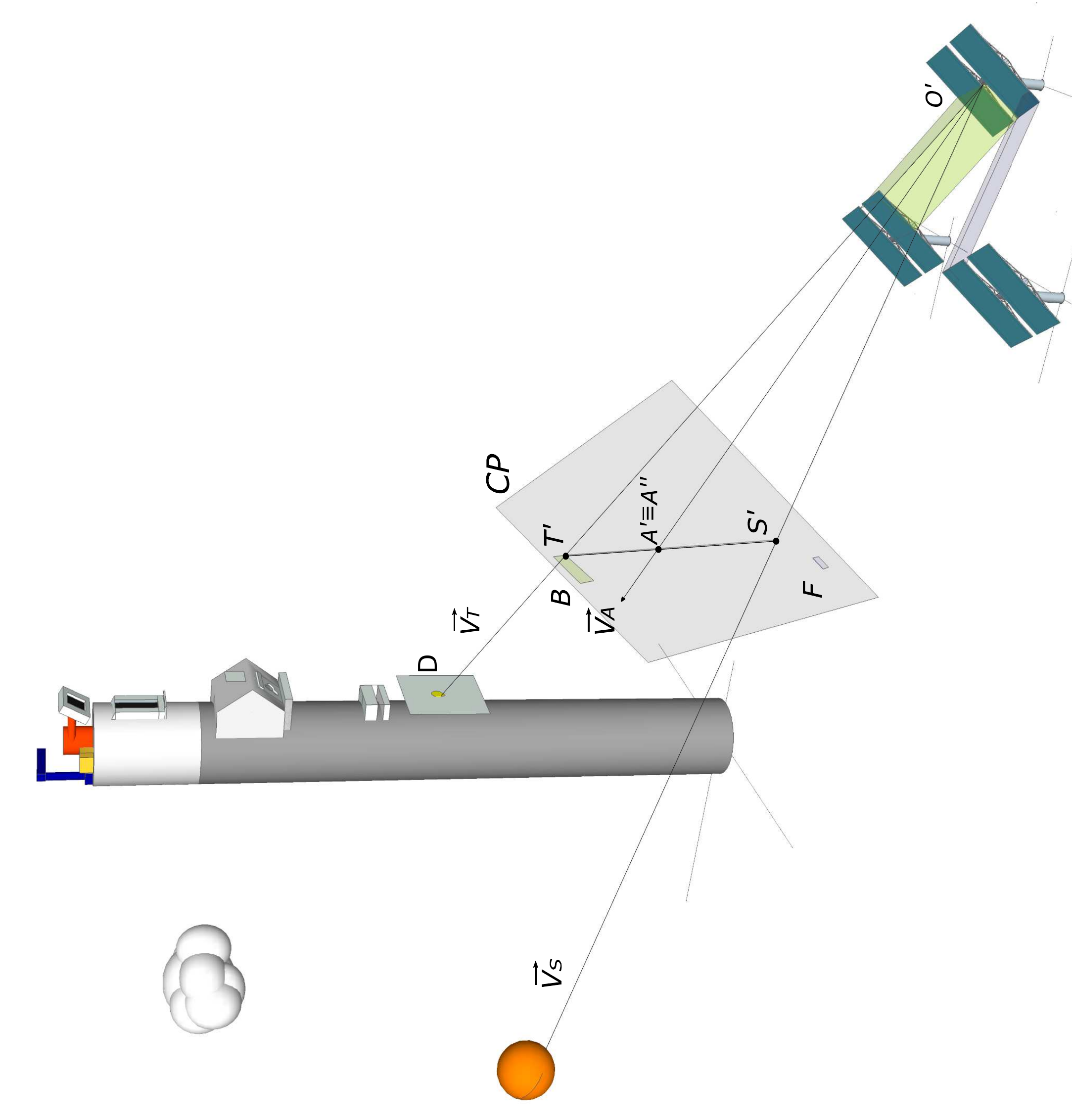}
    \caption{Central tower system}
    \label{f:TowerSystem}
  \end{figure}

As commented before, \emph{STSs} are employed to align in a optimum way the solar collector with the Sun. Thus, for a correct Sun tracking, \emph{STSs} need to know the relative Sun position in the sky, the receiver position and the collector aiming point. Each of these three points together with the point on the origin collector surface closest to the rotation axis ($O'$), form respectively the solar vector ($ \vec{V_S}$), the reflected or target vector ($ \vec{V_T}$) and the aiming vector ($ \vec{V_A}$). On the one hand, these three vectors are usually computed by solar equations, which is a functional method but has several limitations such as time and location dependence, on the other hand, electro optical sensors are only able to obtain $ \vec{V_S}$, limiting the range of application of the system. In this work the new approach, that tries to eliminate these limitations, has been implemented in a central tower system (\rfig{f:TowerSystem}), which can require the most complex \emph{STSs}.

\begin{figure}[h]
   \centering
    \includegraphics[angle =0, width = 0.8\columnwidth]{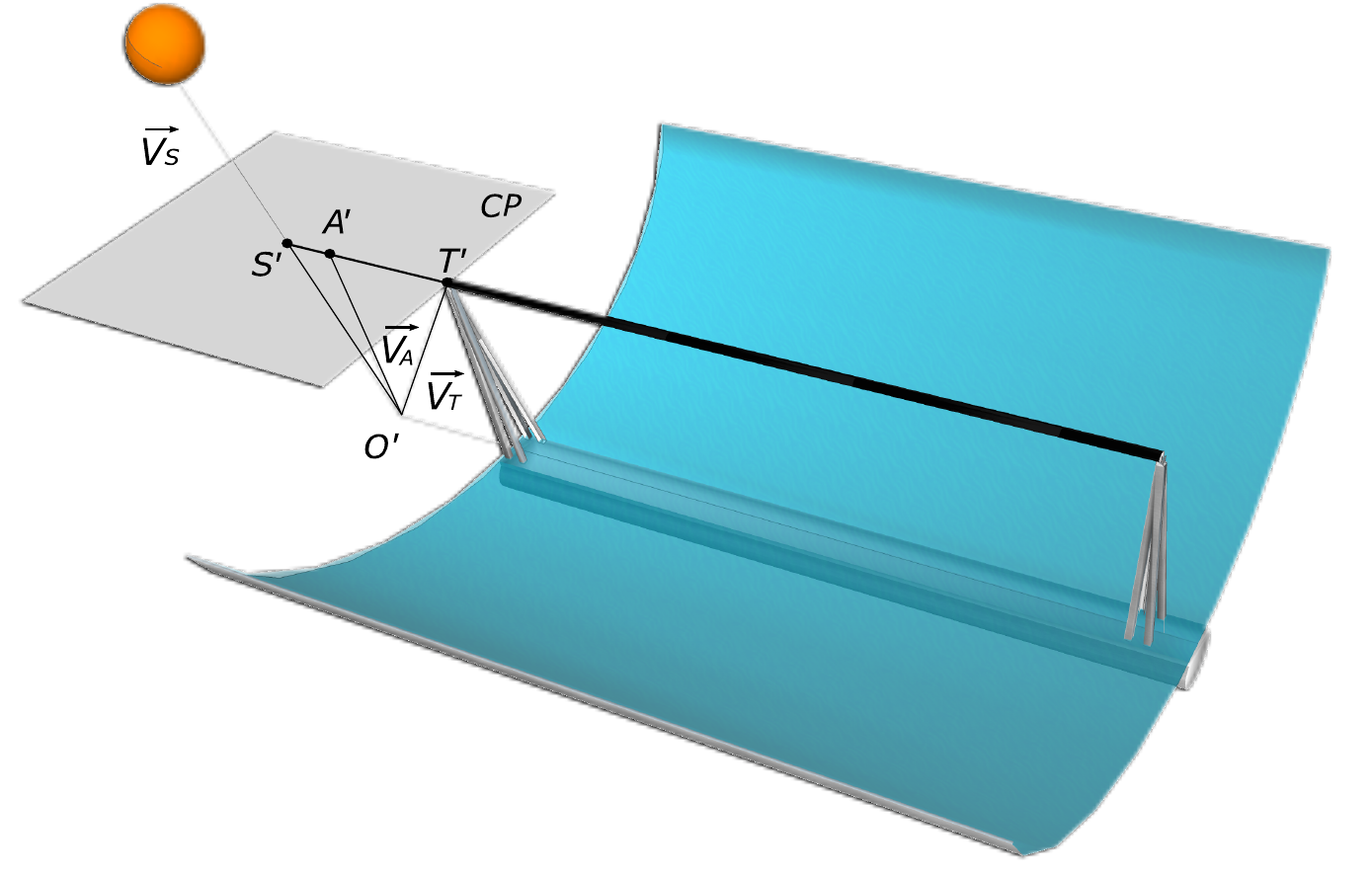}
    \caption{PTC system}
    \label{f:CCPSystem}
  \end{figure}

For that the picture taken by a camera, placed in $O'$ and whose optical axis is arranged parallel to the optical axis of the heliostat, enables detecting $S'$, $A'$ and $T'$, which represent the intersection between the camera plane {($CP$)} and $ \vec{V_S}$, $ \vec{V_A}$ and $ \vec{V_T}$ respectively. With this arrangement, for a  correct heliostat alignment, the vector $ \vec{V_A}$ should intersect the middle point ($A''$) of the segment formed by $S'$ and $T'$ in the $CP$. The differences between $A'$ and $A''$ are known as \emph{tracking error} and it is employed as the main input for the control system. Note that considering some simplifications,  this approach can be applied to any kind of solar collector system like parabolic trough collectors \emph{PTC} ({\rfig{f:CCPSystem}}), parabolic dish or \emph{STS} PV applications. For example, in the \emph{PTC} systems (\rfig{f:CCPSystem}), the Sun tracking problem is reduced to one dimension, so only one of the {\emph{CP}} and key vectors dimension are required. Also, for parabolic dishes or two axis PV trackers, the $ \vec{V_A}$ and $ \vec{V_T}$ are usually the same.
 
Furthermore, this approach enables detecting several key variables related to the control of \emph{STSs} that currently are not considered or computed separately due to the difficulty for obtaining them. For example, Sun  shadow ($F$) projected over the collector or concentrated solar flux blocked ($B$) by other heliostats or objects can be detected by analyzing the image.

Also, clouds detection enables predicting cloud movements, helping in this way to mitigate the effects produced by these transients. 

Another key variable is the concentrated solar flux distribution ($D$) or the total solar flux that reaches the receiver, previous works \cite{Ballestrin2004,Ho2012}  make use of digital images to measure these variables. Also, recently a method for atmospheric extinction measurement based on digital images \cite{Ballestrin2016} has been presented.  Atmospheric extinction of solar radiation flux reflected by the collectors is  an important cause of energy loss in large solar tower plants. Therefore,  the implementation of these methods based on digital images together with the new approach for \emph{STS} based on computer vision, enables obtaining these key variables and develop new control strategies. Finally, the joint use of this new approach and traditional \emph{STS} control techniques can lead to a closed loop control system.

\begin{figure}[h!]
   \centering
    \includegraphics[angle =0, width = 0.7\columnwidth]{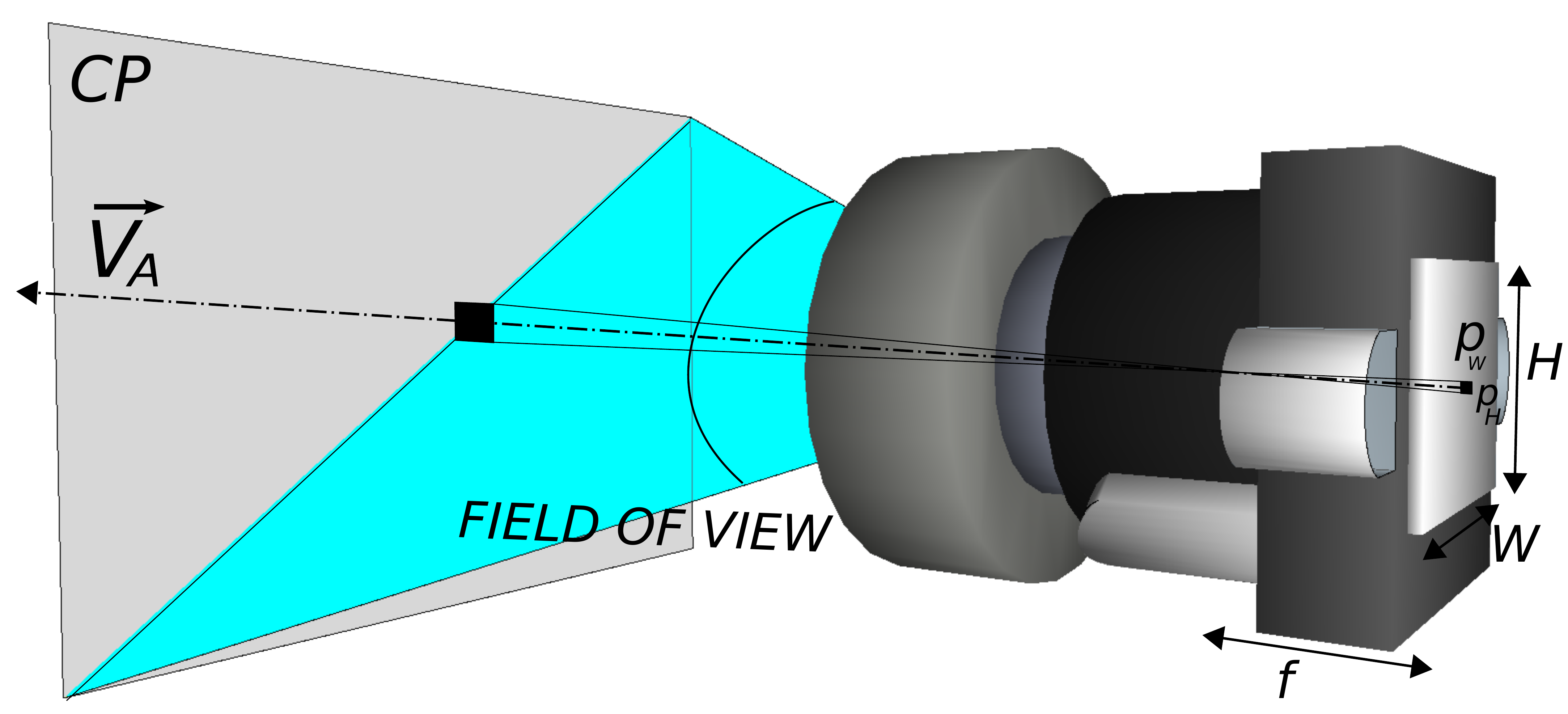}
    \caption{Camera plane, field of view and accuracy.}
    \label{f:cam}
  \end{figure}

In {\emph{STSs}} based on computer vision, different camera settings and parameters (see {\rfig{f:cam}}) should be taken into account since they strongly affect the outcome. The required geometrical parameter, Field Of View ({\emph{FOV}} in {\emph{mrad}}), is determined by the {\emph{STS}} application requirements related to the geometrical layout. For example, tower systems demand a {\emph{FOV}} wider than PV trackers due to the angle between $\vec{V_S}$ and $\vec{V_T}$ is larger. Other relevant geometrical parameters are focal length ({\emph{f}}, measured in {\emph{mm}}) and sensor size, both related to the {\emph{FOV}}. Camera resolution $H*W$ (image height and width in pixels) together with sensor size determine the pixel height and width in $mm$ ($p_W$ and $p_H$), which usually have the same value ($p$).

The uncertainty of the aiming point angle ($U$) for \emph{STSs} based on computer vision can be estimated as the pixel projection size in {\emph{mrad}} over {\emph{CP}} in {\emph{A'}}. It can be computed from {\emph{p}} and {\emph{f}}, see {\requ{eq:uncertainty}}. The minimum camera resolution can be therefore determined from a particular uncertainty.

\begin{equation}
 U = \arctan{\left(\frac{p}{f}\right)}.
 \label{eq:uncertainty}
\end{equation}
  
\subsection{Convolutional Neural Networks}

The new approach is based on computer vision, specifically in deep Convolutional Neuronal Networks {(\emph{CNNs})} for objects localization and detection due to the large number of advantages that it presents, although any other technique based on computer vision could be used. 

Models based on neuronal networks works like regular models, take an input (image), and generate an output (object positions). Neural networks models transform the input through a series of hidden stages (layers), composed by a set of neurons (small and simple processing unit). Each neuron is connected to many others neurons creating a massively parallel processing model. Neurons have learnable parameters (weights and biases), that modify the individual output. 

Weights and biases and other general configuration parameters (such as the learning rate, number of epochs, etc.) must be adjusted in a iterative process called learning or training. The training is an optimization problem, where some parameters are set to minimize a loss function on a training dataset. The loss function expresses the discrepancy between the ground truth and the neural network prediction.

The goal of machine learning is not only to minimize the difference between predictions and the training dataset, but to generalize to be applicable to unseen samples. For that reason, there is commonly an independent dataset used for validation.

A deep neural network is a network with hidden layers between the input and output layers. A \emph{CNN} is a particular class of deep neural network commonly applied to image and video recognition. Convolution networks are inspired by biological processes where the connectivity pattern between neurons resembles the organization of the visual cortex. Among the hidden layers of a \emph{CNN}, there are convolutional layers. A convolutional layer applies a convolution operation to the input, which emulates the response to visual stimuli, and passes the result to the next layer. Convolutional layers have a fixed number of weights determined by the choice of filter size and number of filters, but independent of the input size. This improves performance and reduces memory footprint compared to fully connected layers. During training, a \emph{CNN} learns the filters, which in traditional imaging processing are manually programmed.

Typical object detection networks such as Region-\emph{{CNN}} (\emph{R-CNN}) and \emph{Fast R-CNN} use a region proposal algorithm to select the different regions of interest (\emph{ROIs}) on a picture before running the \emph{CNN} to classify the regions. The difference between them is how \emph{ROIs} are selected to process and how these regions are classified. This work considers  a new approach known as \emph{Faster R-CNN} \cite{Ren2017}, in which a  \emph{CNN} is employed in the region proposal mechanism (Region Proposal Network \emph{RPN}) and a detector based on \emph{Fast R-CNN}  classifies the \emph{ROI}. \emph{RPN} and \emph{Fast R-CNN} can be merged into a unified network by sharing the common set of convolutional layers. This merger causes a problem in networks trainings, because two independent network trainings must be applied to each network, thus each training modifies in different ways the common convolutional layers. To solve this problem, a new training scheme is applied, it is implemented in the Computer Vision System Toolbox in Matlab \cite{MATLAB:2017,TheMathWorksInc.2013}, which alternates between the region proposal and the object detection, producing a unified network with convolutional features that are shared between both tasks. 

A pretrained network called \emph{Alexnet} has been employed in the present work as a starting point to learn the new task because transferring learning is much faster and easier than constructing and training a new network, furthermore the pretrained networks have already learned a rich set of features. \emph{Alexnet} is a \emph{CNN} composed by 25 layers and trained on a subset of the ImageNet database for ImageNet Large-Scale Visual Recognition Challenge (ILSVRC) \cite{Russakovsky2015}. 

\begin{figure*}[h!]
   \centering
    \includegraphics[angle =0,  width = 1\columnwidth]{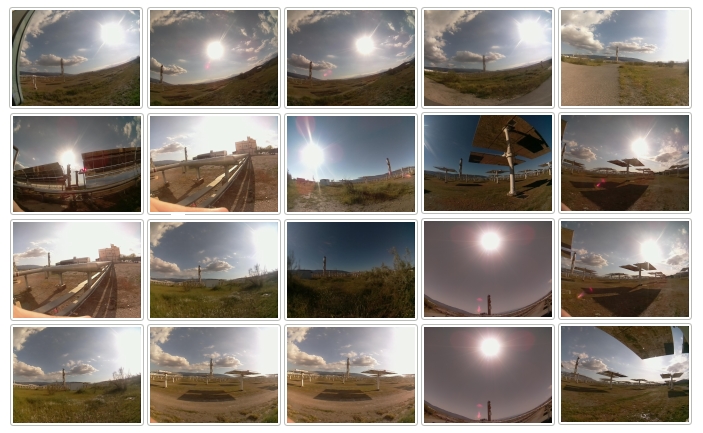}
    \caption{Training image examples.}
    \label{f:Trainingset}
  \end{figure*}

In this work, \emph{Alexnet}  has been retrained with a large training image set (see \rfig{f:Trainingset}) of CESA central tower system  located in Plataforma Solar de Almería, which was taken for this purpose with the open hardware employed for the implementation. Also, the images have been analyzed and the \emph{ROI} labeled according to the new four object classes (heliostat, target, Sun and cloud). 

\begin{figure}[h!]
   \centering
    \includegraphics[angle =0, width = 1\columnwidth]{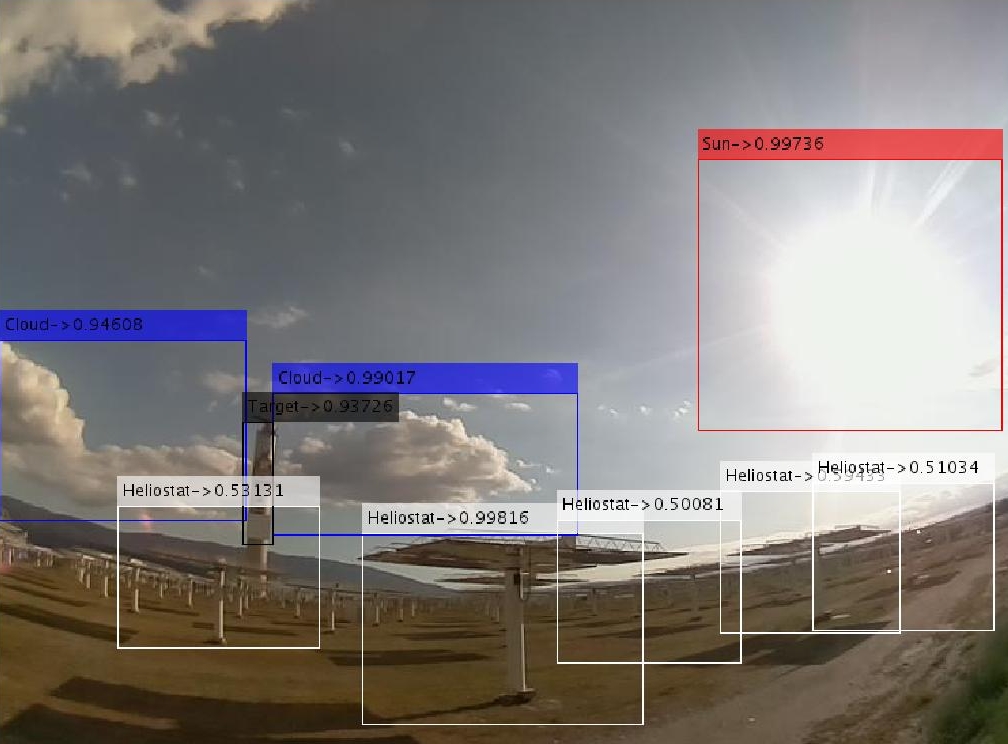}
    \caption{Image analyzed example.}
    \label{f:Analyzedimage}
  \end{figure}

\rfig{f:Analyzedimage} shows one of the image analyzed by the new trained neuronal net, in which the results have been superimposed. In this figure, the red, blue, white and black marked \emph{ROIs} indicate the regions that have been classified according to the new  classes (Sun, cloud, heliostat and target respectively). Each \emph{ROI} has a title with the class name and the score between 0 and 1, larger score values indicate higher confidence in the detection.

Notice that other image processing techniques can be employed for this purpose. For example, Sun center is defined in this work by the segmentation of the Sun \emph{ROI}. In computer vision techniques like segmentation, the image color representation takes a relevant role. The most usual color representation \emph{RGB} is not suitable because \emph{RGB} representation is affected by the bright level. Another color representation called \emph{HSL} defines a color in terms of saturation (the amount of chroma), its constituent components hue (the angle on the color wheel) and lightness (how bright the color is). In \emph{HSL}, brightness can be separated to make the colors less influential to the light intensity impact improving the discrimination of the object \cite{Lee2013}. For this reason, the \emph{HSL} representation is employed in this proposal to determine the position of the Sun's center in the Sun \emph{ROI}.

\subsection{Low cost hardware}

Nowadays in solar industry, the collector fixed costs are one of the factors which determine the optimum collector size. For central tower systems  the  optimum heliostat surface size is  $150$ $m^2$. The cost of the control electronic devices employed in \emph{STSs} like processors, encoders, limit switches, junction boxes, connectors, wiring, remain constant at $\$1000-3000$, although the collector size changes \cite{Kolb2007,Blackmon2013}. 

The proposed approach in this work can simplify the system and reduce the number of components. Furthermore, in the implementation of the approach a low cost hardware platform called \emph{Raspberry Pi} \cite{pi2013raspberry} has been employed. This reduces the fixed cost associate to the \emph{STS} control hardware from $\$290$ to less than $\$75$ approximately, removing  costs in electronic devices such as limit switches, GPS or encoders that are not needed. Communication  wire costs can be removed due to the great connectivity offered by this type of platforms, in this case Wifi and bluetooth. Also, the new hardware can be powered with a small pv cell and a battery. For all that, this new approach can make \emph{STSs} more flexible, autonomous \cite{garcia2002first}, easy to use, cheap to manufacture and install, which leads to new cost reductions. Furthermore, cost reduction in fixed control costs results in direct savings and also enables the reduction of the optimal heliostat area with additional cost reductions in the cost per unit area due to the dependence on imposed loads \cite{Blackmon2013}.  Finally, this hardware facilitates the implementation of the Internet Of Thing (\emph{IOT}) concept in solar energy and by extension, the integration of this type of technologies in the new energy distribution paradigm called smart grids \cite{REALCALVO201756}.

\begin{figure}[h!]
   \centering
    \includegraphics[angle =0, width = 0.9\columnwidth]{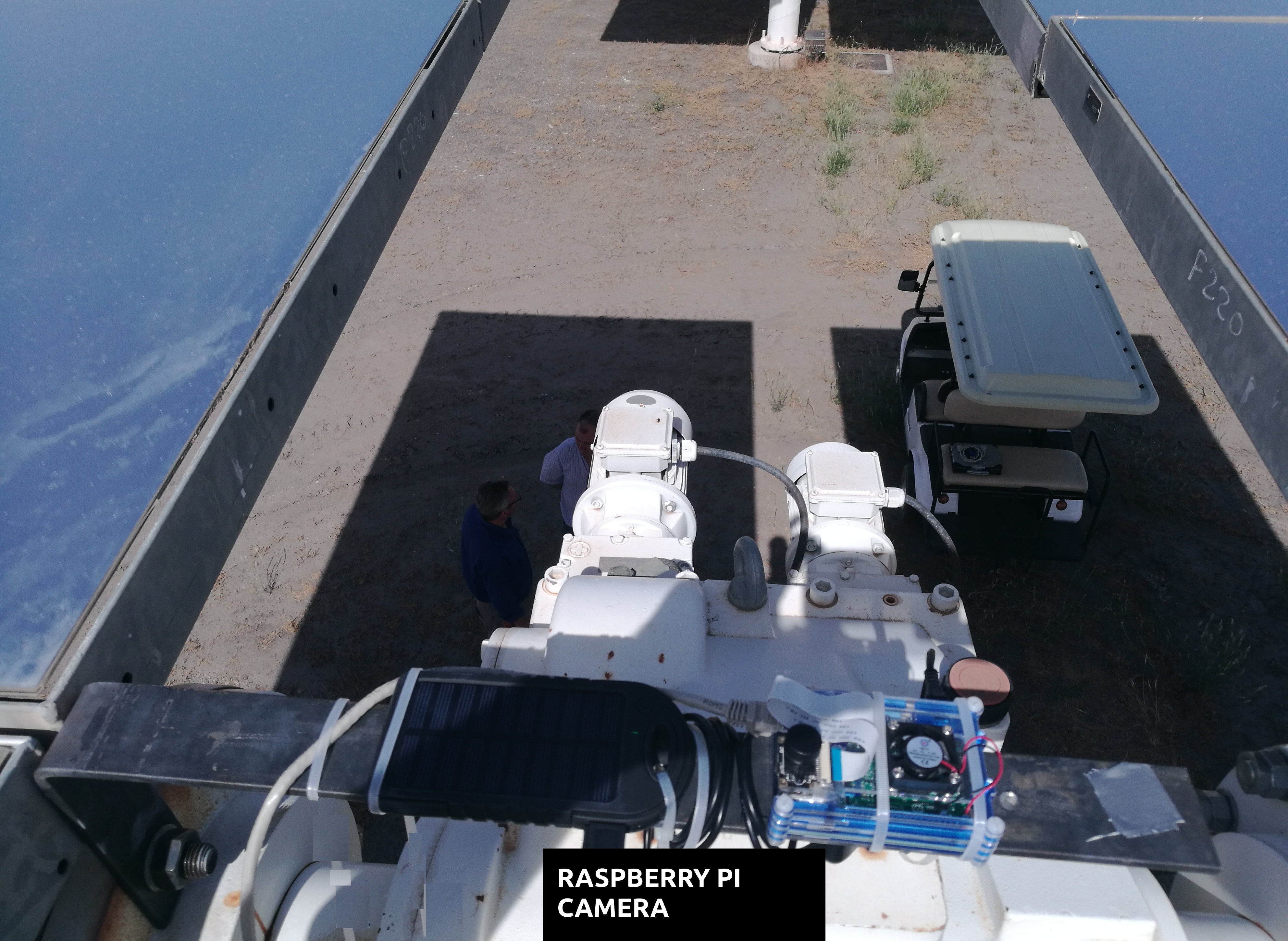}
    \caption{Hardware assembly.}
    \label{f:Prototype}
  \end{figure}

The hardware employed (\rfig{f:Prototype}) was a \emph{Raspberry Pi 3} together with a \emph{Raspicam}. \emph{Raspberry Pi 3} is a low cost credit-card-sized computer that compared with  industrial hardware is cheaper  while being fully functional and adding more capabilities. It can be found a large number of platforms like \emph{Raspberry Pi} in the market that can be used for implementing this approach, most of them enable low cost replacements. The popularity of these hardware platform offers a lot of hardware extensions, generally low cost, which may improve the \emph{STS}. The software employed in this work has been \emph{Computer Vision System Toolbox} \cite{TheMathWorksInc.2013,Ren2017} and \emph{MATLAB Support Package for Raspberry Pi Hardware}, although the great adaptability of the hardware enabling using other software packages.

\section{Test}
 
\begin{figure}[h!]
   \centering
    \includegraphics[angle =0, width = 1\columnwidth]{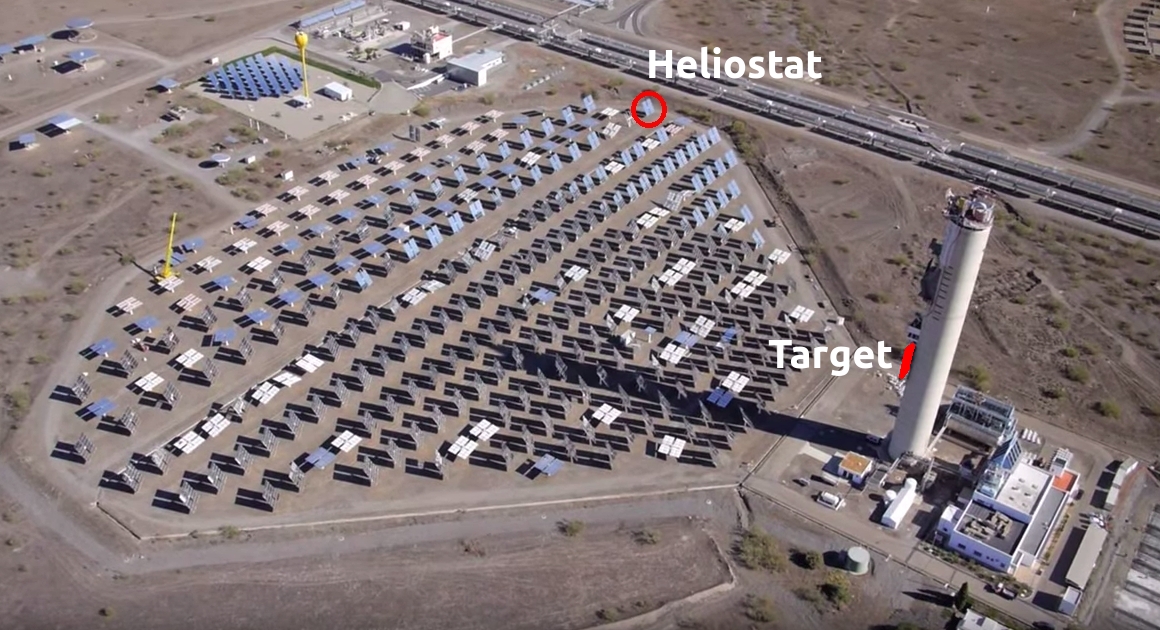}
    \caption{Cesa scheme.}
    \label{f:Figuracampo}
  \end{figure}

In order to test the new approach described above, the hardware was fixed to one of the heliostat of the CESA tower central system at \emph{PSA} (\rfig{f:Prototype}) and two new image sets were taken while the heliostat remained under the control of the traditional control system. The heliostat selected (\rfig{f:Figuracampo}) is located in the eastern end of the \emph{CESA} field, so it is one of the most distant from the white target. The heliostat solar tracking uncertainty with the traditional  Supervisory Control And Data Acquisition (\emph{SCADA}) is $1.2$ {\emph{mrad}} for each axis.

Also with the configuration of the camera employed during the test (camera resolution $800 x 600$ {\emph{px}}, wide angle {\emph{FOV}} lens $150^{\circ}$, camera sensor size $3.76 x 2.74$ {\emph{mm}}, {\emph{f}} $2.35$ {\emph{mm}} {\cite{pi2013raspberry}}), the solar tracking uncertainty of the new approach based on {\emph{CNN}} according to {\requ{eq:uncertainty}} is $2$ {\emph{mrad}}. Note that with this configuration, a camera resolution of $1600 x 1200$ px, would reduce the uncertainty below $1$ {\emph{mrad}}.

  \begin{figure}[h!]
   \centering
    \includegraphics[angle =0,width = 0.8\columnwidth]{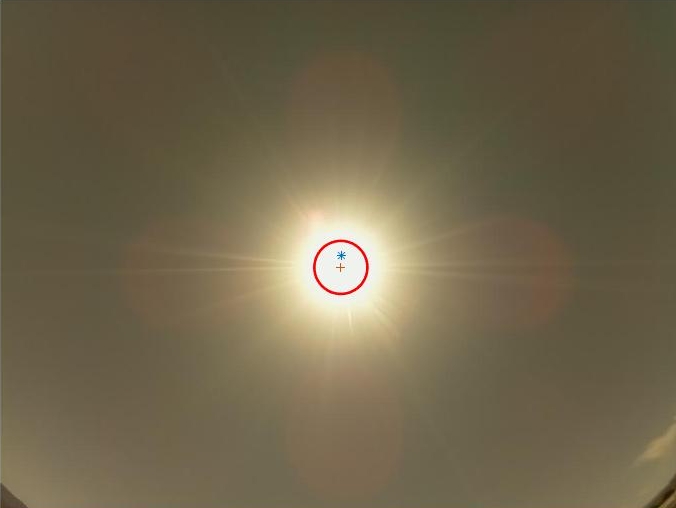}
    \caption{Aiming error.}
    \label{f:AimingError}
  \end{figure} 
  
The first image set was taken while the heliostat pointed directly at the Sun in tracking mode, with this configuration  the mean differences between $ \vec{V_a}$ and $ \vec{V_s}$ during the test can be considered as the constant   \emph{aiming error} and can be corrected in the following analyses. \rfig{f:AimingError} shows one of these pictures analyzed and the differences in pixels between  $ \vec{V_s}$ (red cross) and  $ \vec{V_a}$ (blue star).
  
\begin{figure*}[h!]
   \centering
    \includegraphics[angle =0, width = 1\columnwidth]{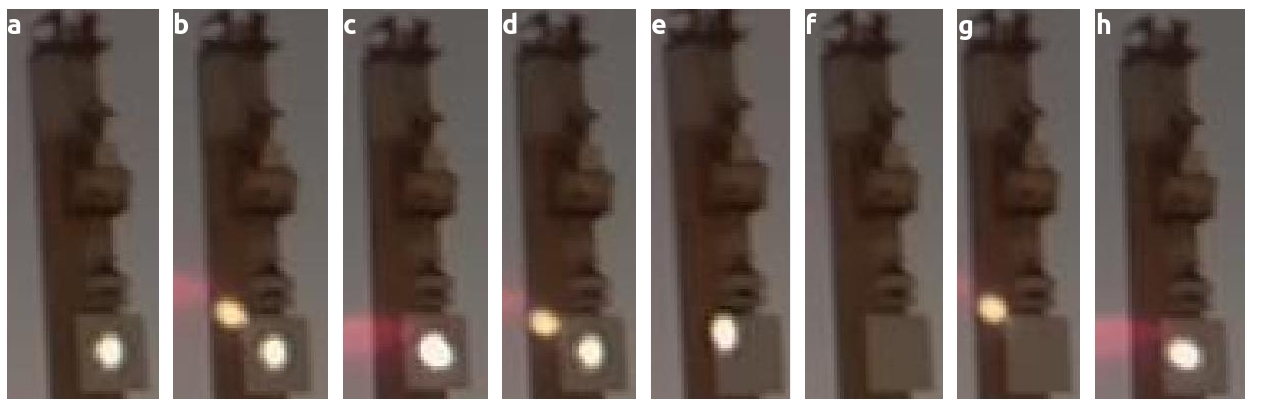}
    \caption{Test captions.}
    \label{f:targettest}
  \end{figure*}

The second image set was taken in tracking mode where the white target is the aiming objective. \rfig{f:targettest}.\emph{a} shows the initial instant,  where the white target is being focused by an heliostat and the modified heliostat (heliostat with new hardware) is also sent to focus at the white target. In \rfig{f:targettest}.b, the concentrated  solar distribution due to the modified heliostat is in transition through the upper corner to the center of the white target. Next, the concentrated solar distribution of each overlaps in the center of the white target (\rfig{f:targettest}.c). After that, the first heliostat was sent out of the white target (\rfig{f:targettest}.d), remaining the modified heliostat. The last pictures in \rfig{f:targettest} show how the modified heliostat was sent out and then sent back to the white target center. The data computed by the traditional \emph{STS} of the modified heliostat was in turn stored for further analysis with the main goal to compare with the Convolutional Neural Networks object detector results. 

\section{Results}

\begin{figure*}[h!]
   \centering
    \includegraphics[angle = 0, width = 1\columnwidth]{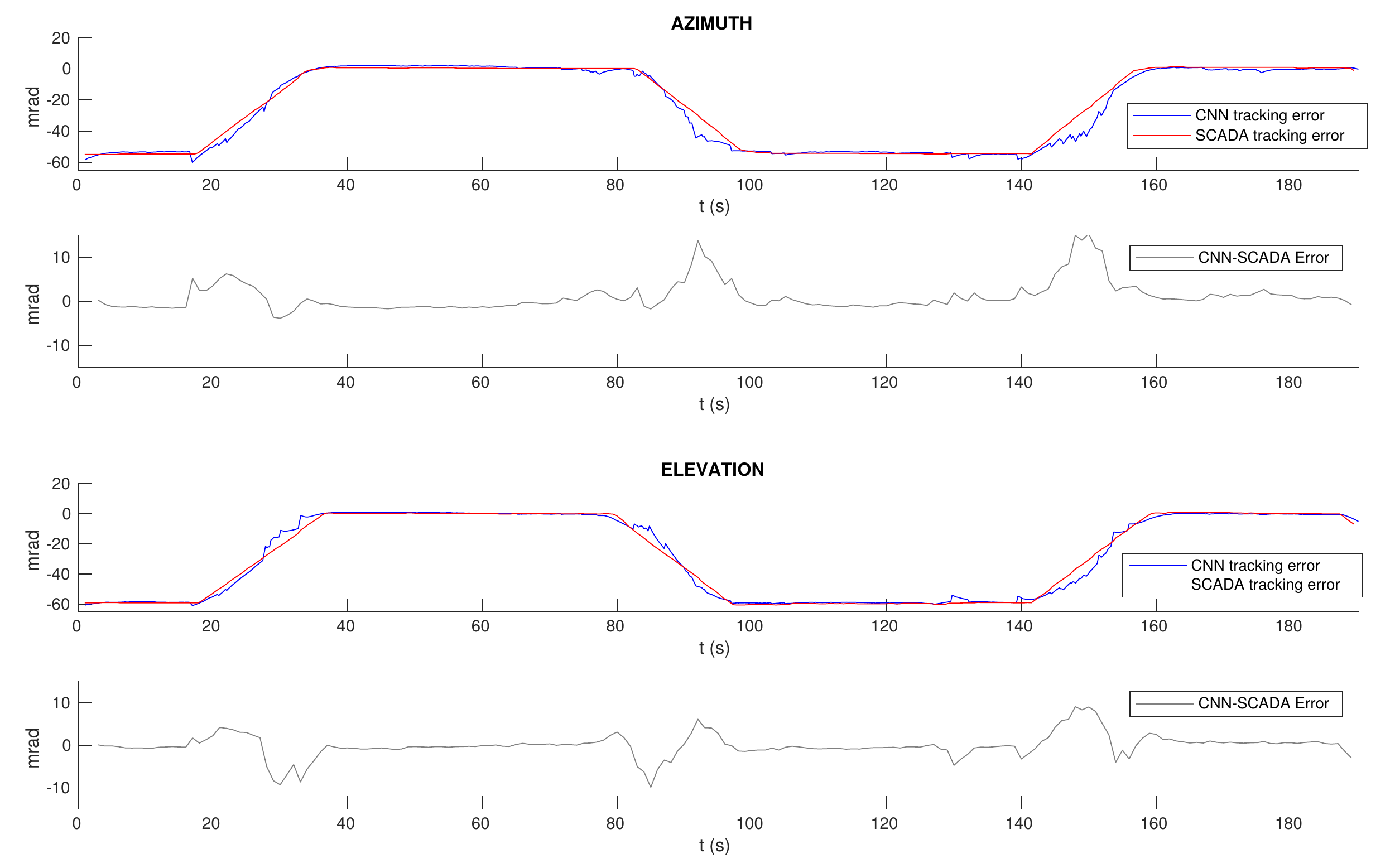}
    \caption{Tacking signals and error evolution.}
    \label{f:trackingerror}
  \end{figure*}

The previous image sets were analyzed with an algorithm developed for this purpose. The algorithm analyzes the images one by one by means of the Convolutional Neural Networks object detector to find all the classes to develop the Sun tracking task properly, as well as to find key parameters such as clouds or shadows and blocks originated by other heliostats. Results due to the analysis of the last image set are shown in \rfig{f:trackingerror}. Blue and red lines represent the tracking error obtained through the analysis of the Convolutional Neural Networks object detector and data stored in the {\emph{SCADA}} of the modified heliostat. The grey line represents the difference between both signals (\emph{CNN-SCADA} Error).

It can be noted that the tracking error measured by the two methods (\emph{CNN} object detector and traditional \emph{SCADA}) behave in the same way for both axes, the widest differences (15 mrad) are seen in large transitions when the heliostat enters or leaves the tracking mode. Therefore, although they are important, these differences are not the most relevant for the heliostat performance. In these periods, the \emph{CNN-SCADA} error patterns suggest that they could be caused by the deformation of the structure during large movements, when the heliostat rotates at the highest speed. Note that the {\emph{STS}} based on {\emph{CNN}} is less affected by structural deformations due to its position is directly determined by the concentrator optical axis, and not from the rotation axes.

The differences kept below 3 mrad for steady-state and narrow movements, besides at 130 s and 140 s time instants, when the error increased up to 4 mrad due to an inaccurate Sun detection caused by an incorrect camera set up.

The tracking error is measured by Convolutional Neural Networks object detector and traditional \emph{STS} \emph{SCADA} in \emph{pixel} and \emph{mrad} respectively, but in \rfig{f:trackingerror} both are shown in the same unit. \emph{Pixel} and \emph{mrad} are related by the camera sensor resolution, {\emph{FOV}} and the geometry of the system (slant range and relative position), although the objective of the new approach is to perform the control directly with the error measured in \emph{pixel} to avoid the individual configuration of the \emph{STS} in a large system.

Nevertheless, the method proposed in this paper provides similar error values to the traditional method, with an error within the acceptable range for most applications. Better Convolutional Neural Networks object detector training and higher resolution cameras would significantly reduce this difference with the traditional system. The use of camera filters and fine-tuning of the camera setting may improve the performance to a higher level of precision.

Furthermore, it can be noted that the operation of other systems that jointly use certain parts of the system (white target), does not disturb the normal operation of the \emph{STS} proposed.

\section{Conclusions}

According to the results obtained in the tests, it can be concluded that the new approach proposed for \emph{STSs} is valid, fully functional and  shows a wide margin for improvement. The new approach is independent of solar technology, system size, location and time. It is not affected by \emph{aiming errors} like pedestal tilt, wind loads or apparent Sun position. Furthermore, the proposed approach provides advantages such as the ability of cloud, block and shadow detection, atmospheric attenuation or concentrated solar radiation measurement,  which can improve the control strategies of the system and therefore the system performance. The new approach combined with traditional control techniques makes possible closed-loop control schemes.

In addition, provide the system with a versatile and cheap microprocessor, enables to reduce costs related to the \emph{STS} and other aspects such as communication or first start up and gives greater autonomy and flexibility.

Future work includes testing these methods and algorithms and performing extensive training of the neural network in order to further improve the obtained results while reduce the computational cost. Another important task is the autonomous control of an heliostat using the new presented tracking method.

\section{Acknowledgments}
This work has been funded by the National R+D+i Plan Project DPI2014-56364-C2-2-R of the Spanish Ministry of Economy, Industry and Competitiveness and ERDF funds.
\clearpage
\section*{References}

\bibliography{mybibfile}

\end{document}